\documentclass{article}
\usepackage{ijcai22}

\usepackage{times}
\usepackage{soul}
\usepackage{url}
\usepackage[hidelinks]{hyperref}
\usepackage[utf8]{inputenc}
\usepackage[small]{caption}
\usepackage{graphicx}
\usepackage{amsmath}
\usepackage{amsthm}
\usepackage{booktabs}
\usepackage{algorithm}
\usepackage{algorithmic}
\usepackage{tikz}
\usepackage[caption=false]{subfig}
\usepackage[normalem]{ulem}
\usepackage{algorithm}
\usepackage{array}
\usepackage{color}
\usepackage{amsmath,amssymb}
\usepackage{multicol}
\usepackage{multirow}
\usepackage{array}
\usepackage{bm}
\usepackage{mathtools}
\usepackage{amssymb}
\usepackage{url}
\usepackage{appendix}

\title{Table Pre-training: A Survey on
Model Architectures, Pre-training Objectives, and Downstream Tasks}

\author{Haoyu Dong\textsuperscript{1},
Zhoujun Cheng\textsuperscript{2},
Xinyi He\textsuperscript{3},
Mengyu Zhou\textsuperscript{1},
Anda Zhou\textsuperscript{4},
Fan Zhou\textsuperscript{2},
Ao Liu\textsuperscript{5},\\
Shi Han\textsuperscript{1},
Dongmei Zhang\textsuperscript{1}\\
\{hadong, mezho, shihan, dongmeiz\}@microsoft.com,
\{blankcheng, zhoufan98\}@sjtu.edu.cn,
hxyhxy@stu.xjtu.edu.cn,
a.d.zhou@sms.ed.ac.uk,
zeitmond@gmail.com}

\begin{document}
\maketitle

\begin{abstract}
Following the success of pre-training paradigm in the natural language domain, a flurry of table pre-training frameworks have been proposed and have achieved new state-of-the-arts on various downstream tasks such as table question answering, table type recognition, column relation classification, table search, and formula prediction. 
Various model architectures have been explored to best leverage the characteristics of structured tables, especially specially-designed attention mechanisms.
Moreover, to fully exploit the supervision signals in unlabeled tables, diverse pre-training objectives have been designed and evaluated, for example, denoising cell values, predicting numerical relationships, and learning a neural SQL executor. This survey aims to provide a review of model designs, pre-training objectives, and downstream tasks for table pre-training, and we further share our thoughts on existing challenges and future opportunities.
\end{abstract}

\section{Introduction} \label{sec:intro}

Tables are widely used to organize and present data in various document types and database systems such as webpages, spreadsheets, PDFs, and MySQL, gaining increasing attention from the research community. 
Following the success of large-scale pre-training in natural language (NL), a flurry of research works have been proposed to leverage unlabeled tables for self-supervised pre-training and achieve promising results in table type classification~\cite{wang2021tuta}, cell type classification~\cite{gol2019tabular,wang2020tuta}, table question answering (QA)~\cite{herzig2020tapas,yin2020tabert}, table search~\cite{wang2021retrieving}, entity linking~\cite{deng2020turl}, column type identification ~\cite{chen2019colnet,guo2020web}, table augmentation~\cite{deng2020turl,iida2021tabbie}, formula prediction~\cite{cheng2021fortap}, etc. 
On the one hand, similar to NL that has already proved the success of large-scale pre-training, tables have dense semantics stored in textual headers, captions, and notes.
On the other hand, different from NL, tables have distinct information (intuitive formats, well-organised numerical values, formulas, etc.) and various structures (relational tables, entity tables, matrix tables, forms, etc.), and thus require special model designs and pre-training objectives to achieve optimal results.

\begin{figure}[th]
\begin{center}
\includegraphics[width=3.4in]{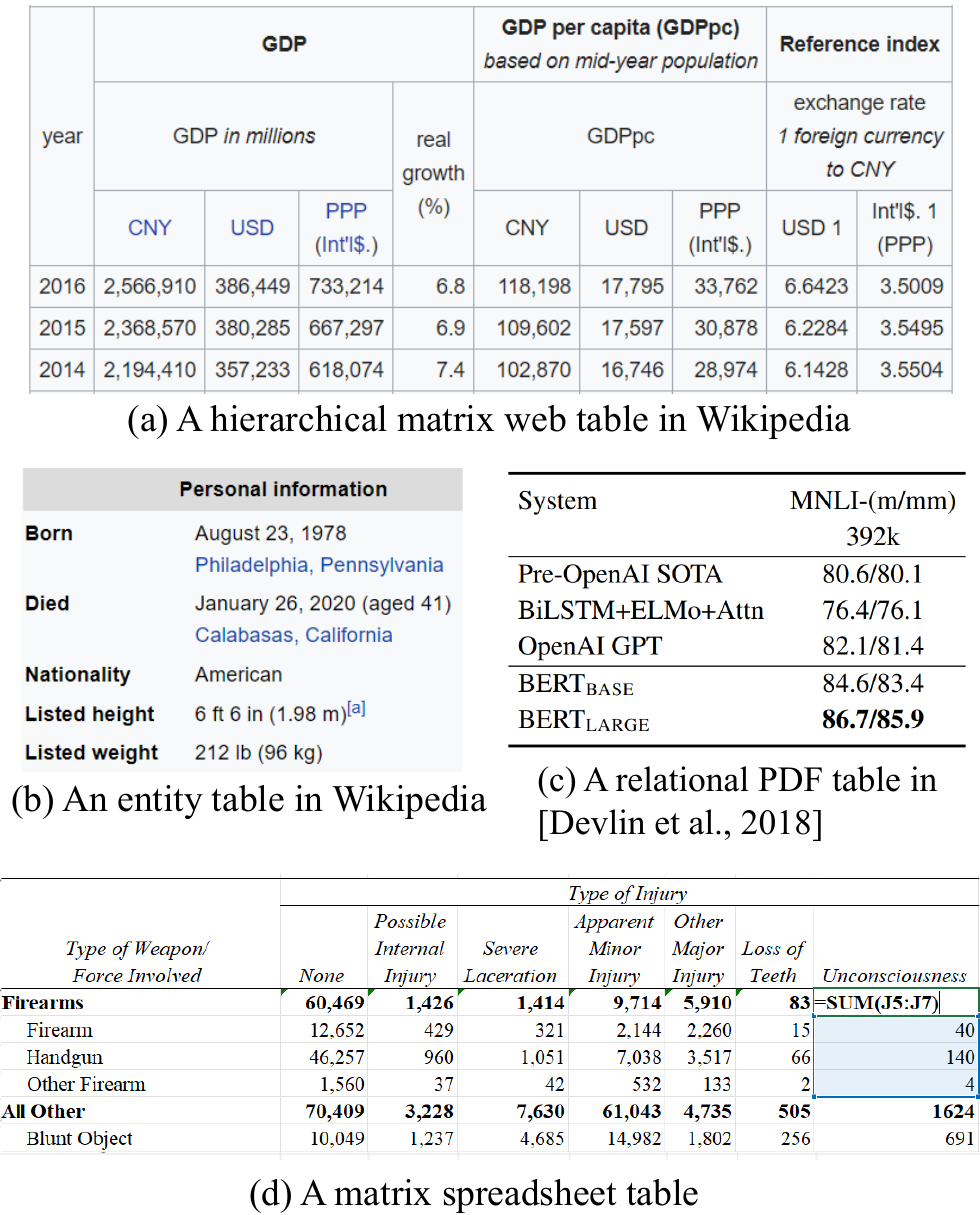}
\end{center}
\caption{Examples of real-world web, PDF, and spreadsheet tables. }
\label{fig:realtable}
\end{figure}

To best leverage table characteristics while maintaining capabilities to understand text within/out of tables, various Tabular Language Models (TaLMs) are proposed for table pre-training. For example, TaBERT~\cite{yin2020tabert} encoded tables and text by concatenating a row-wise transformer with column-wise vertical attention layers, pre-trained with Masked Language Models (MLM) and Masked Column Prediction (MCP), and achieved SOTA results on benchmarks of table QA. TaPas~\cite{herzig2020tapas} pioneeringly proposed an end-to-end table-text joint reasoning framework using transformers without explicitly generating logical forms for table QA, and TaPas 
also employed MLM for pre-training.
TURL~\cite{deng2020turl} was the first work to learn entity representations from relational tables to enhance table knowledge matching and table augmentation, and it restricted each cell to aggregating information from the located row and column via masked attention.
TUTA~\cite{wang2020tuta} then extended the success of table pre-training to generally structured tables using tree-based attention and tree-based positional encoding based on a novel unified bi-tree structure. TUTA achieved new SOTA results on five representative table structure understanding datasets.
Far different from previous pre-training objectives, TaPEx~\cite{liu2021tapex} explored to learn a neural SQL executor and demonstrated surprising effectiveness on table-text joint reasoning. Recently, UnifiedSKG~\cite{xie2022unifiedskg} explored to directly fine-tune T5 on 21 datasets across 6 tasks and achieved promising and even SOTA results.

We believe that the structured table provides a distinct perspective to explore frontier neural architectures and inspire new research directions. Since the table often interacts with programming languages such as SQLs and spreadsheet formulas, it additionally spawns cross-domain applications such as semantic parsing~\cite{yu2018spider}, logic-to-text~\cite{chen2020logic2text}, and formula prediction~\cite{chen2021spreadsheetcoder}.
In this paper, we first introduce preliminaries of table types, table structures, cell information, and table corpora in Section~\ref{sec:pre}. Then we give a comprehensive review of table modeling architectures, table pre-training objectives, and downstream tasks in Sections~\ref{sec:model},\ref{sec:pretraining objectives},\ref{sec:task}. At last, we share our vision on table pre-training in Section~\ref{sec:conclusion}. 

\section{Preliminaries}
\label{sec:pre}
\label{sec:pre-type}
Tables can be roughly categorized into three primary forms: well-structured tables, semi-structured tables, and unstructured tables. 
Database tables are \textbf{well structured} with an ordinarily-defined relational schema so that precise support execution of formal languages such as SQL and R. In contrast, \textbf{semi-structured} tables are usually human-crafted with markup languages or end-user tools such as HTML code, Latex code, spreadsheets, and Word documents. They have flexible structures but lack meta-information to record them and thus challenge precise and automatic information retrieval. Image tables are even \textbf{unstructured} because they only record raw RGB information, e.g., scanned tables from books, screenshots of web tables, and even handwritten drafts. They need to be digitized before any higher-level information retrieval.

In this paper, we mainly focus on semi- and well-structured tables. We \textbf{neglect image tables} because they have distinct visual challenges and are desirable to be discussed separately.

\subsection{Table Structure}
Tables are flexible with various structures, including relational tables, entity tables, matrix tables, layout tables, forms, etc. They also have orientations (horizontal/vertical) and hierarchies (flat/hierarchical). As shown in Figure~\ref{fig:realtable}, the structure of a flat relational table is definite and straightforward in a database-like form, wherein each row is a record, each column is a field, and there is no hierarchy. An entity table simply records an entity and its attributes. 
A matrix table has both horizontal and vertical orientations. In fact, there are various categorization methodologies on the table structure, which are summarized in detail by ~\cite{zhang2020web}.

\subsection{Cell Information}
Typically, a cell is the intersection of one row and one column in a table. And multiple cells can be merged into a larger cell that occupies multiple rows and columns. Cells are basic units to record text, numerical values, formats, formulas, etc.

1)  Text. Text is a critical component in the table to record meta information in headers, notes, and captions, as well as data region cells. Texts in tables are basically in NL but often have short lengths and concise meanings to meet the space restriction in documents.

2)	Numerical values. A large proportion of cells store numerical values. Unlike text, numerical values could have arithmetical relationships, such as sum and proportion, and statistical characteristics, such as distribution and trend.

3)	Visual formats. Tables have various intuitive formats to present the table structure or content, such as border, alignment, background color, and font~\cite{dong2020neural}.

4)	Formulas. In several popular end-user tools such as Excel and Google Sheet, spreadsheet formulas are used to store the logical and numerical relationships between cells.

5) Other elements such as hyperlinks, images, and icons can also be inserted into a cell. A table can even be nested in Word documents by inserting sub-tables into individual cells.

\subsection{Key Differences Between Table and Text}
\label{sec:nlrelation}
Although leveraging frontier LMs for table pre-training is common, we summarize three differences that need to be considered carefully.
(1)	Structures. Unlike sequential NL text, tables have two-dimensional structures with optional top and left headers.
(2)	Visual Formats. Unlike plain NL text, tables usually use intuitive formats to better represent table structures and data highlights, e.g., color, alignment, font.
(3)	Numerical values. Tables usually contain well-organized numerical values for easy-lookup and comparison, and there are various calculation relationships among numerical values, which can be recorded by the spreadsheet formula.

\subsection{Existing Large Table Corpus}
\hspace{\parindent}\textbf{Web Tables}
Large corpora include WDC Web Table Corpus~\footnote{http://webdatacommons.org/framework/} (233M tables), Dresden Web Tables Corpus~\cite{eberius2015building} (174M tables), WebTables~\cite{cafarella2008uncovering} (154M tables), and WikiTables (1.6M tables). More details are summarized by ~\cite{zhang2020web}.

 \textbf{Spreadsheet Tables} FUSE~\cite{barik2015fuse} included 249,376 web-crawled spreadsheets. \cite{chen2013automatic} obtained 410,554 Excel files from 51,252 Internet domains TUTA~\cite{wang2021tuta} collected 13.5 million spreadsheet files from 1.75 million web sites. 

\textbf{CSV Tables}
GitTables~\cite{hulsebos2021gittables} collected 1M+ tables from CSV files in GitHub repositories. Tables in GitTables are found to resemble typical database tables.

 \textbf{Other Kinds of Tables}
TableArXiv~\footnote{http://boston.lti.cs.cmu.edu/eager/table-arxiv/}
contains 341,573 tables extracted from scientific publications on arxiv.org.

\section{Model} \label{sec:model}

Since two-dimensional structure information is crucial for understanding semi-structured tables, many neural architectures have been proposed to jointly capture structure and semantic information.
Table2Vec~\cite{zhang2019table2vec} adopted skip-gram neural network models to train word embeddings with row/column population. 
CNNs~\cite{dong2019semantic,chen2019colnet} were adapted to capture spatial information in two-dimensional tables.
Bidirectional RNNs and LSTMs have been widely used to capture the order of rows/columns~\cite{nishida2017understanding,gol2019tabular,fetahu2019tablenet,kardas2020axcell}. 
Later works explored graph neural networks for table understanding and question answering~\cite{zayats2021representations,zhang2020graph,koci2018table,du2021tabularnet}.
While CNNs, RNNs, and GNNs have been widely used for table modeling, few people adopted them for large-scale table pre-training (except a CNN-based TCN~\cite{wang2021tcn}). The main reason is that \textbf{most of them train (or directly consume) token/word embeddings separately from the CNN/RNN/GNN model}, thus restricting models' capabilities in understanding cell texts together with table structures.

Recently, a flurry of works explored to use transformer-based language models (LMs) for table pre-training, and we call them Tabular Language Models (TaLMs), e.g., TaPas~\cite{herzig2020tapas}, TaBERT~\cite{yin2020tabert}, TURL~\cite{deng2020turl}, TUTA~\cite{wang2021tuta}, Tabnet~\cite{arik2019tabnet}, VIME~\cite{yoon2020vime}, KGPT~\cite{wang2021retrieving}, RPT~\cite{tang2020rpt}, StruG~\cite{deng2020structure},
TabTransformer~\cite{huang2020tabtransformer}, GraPPa~\cite{yu2020grappa}, GAP~\cite{shi2021learning},
BRIDGE~\cite{lin2020bridging},
TABBIE~\cite{iida2021tabbie}, TaPEx~\cite{liu2021tapex}, ForTaP~\cite{cheng2021fortap}, MATE~\cite{eisenschlos2021mate}, STTP~\cite{xing2021sttp}, GTR~\cite{wang2021retrieving}, TableFormer~\cite{yang2022tableformer}, and FLAP~\cite{flap2021}. 
The advantage of using transformers is that they can pre-train semantic and structural representations jointly and inherit the text understanding power of existing NL pre-trained models such as BERT, BART~\cite{lewis2020bart}, and T5~\cite{2020t5}.
Works like UnifiedSKG~\cite{xie2022unifiedskg} and TableGPT~\cite{gong2020tablegpt}, while without table-specific pre-training, directly fine-tuned LMs on table tasks and achieved promising and even SOTA results, demonstrating that \textbf{pre-training is transferable from text to tables}, e.g., in linguistic and world knowledge aspects.
Considering that using transformer-based TaLMs is a common choice for table pre-training, \textbf{in the following subsections, we dig deeper into TaLMs} on sequence serialization, input embedding, the encoder and decoder architecture, the attention mechanism, and model efficiency.

\subsection{Tabular Sequence Serialization}
\label{sec:3.2.1}
TaLMs require a sequence of tokens to be the model input like LMs.
A straightforward yet common method is to linearize raw tables row by row. Most works such as TaPas, MATE, TableFormer, TUTA, and TURL, performed in this way. In TaPEx, the table is also linearized row by row but it additionally inserts several special tokens to indicate table components such as [HEAD] and [ROW] representing the region of table headers and rows respectively. 
TABBIE linearized tables by rows and columns separately for two transformers.y
TableGPT distinctly adapted a template-based table serialization way on relatively simple tables.
Experiments conducted by UnifiedSKG showed that putting external text (like questions) ahead of tables could help T5 to generalize better on tabular tasks.
~\cite{li2021markuplm} directly encoded markup languages like NL. Some works linearized a specific part of a table, e.g., TaBERT \cite{yin2020tabert} linearized most relevant rows to the input utterance, and StruG and GraPPa only took headers as input without data cells. 

\subsection{Input Featurization and Embedding}
\hspace{\parindent}\textbf{Cell Text Encoding}
Most table pre-training methods tokenized cell text using WordPiece and learned token embeddings~\cite{devlin2018bert}, such as TaBERT, TaPas, MATE, StruG, TableFormer, TUTA, ForTaP, and TABBIE. Some works also used the BPE tokenization following Roberta~\cite{liu2019roberta} or BART~\cite{lewis2020bart} , e.g., GraPPa and TaPEx. TURL was initialized by TinyBERT~\cite{jiao-etal-2020-tinybert} and additionally learned embeddings based on an entity vocabulary. Rather than directly using the vocabulary parsed from NL corpora, TUTA constructed a table-specific vocabulary using WordPiece based on large table corpora and merged it with BERT's vocabulary. 

\textbf{Positional Encoding} 
Following NL pre-trained models, most TaLMs embedded 1D sequential positions in the serialized tabular sequence, e.g., TaPas, MATE, StruG, GraPPa, and TaPEx. Some other works divided the whole sequence into multiple pieces and counted positions in each piece separately: TUTA treated each cell as an independent piece and locally encoded positional information of tokens inside a cell; TURL regarded the table caption and the header as two separate pieces. Therefore it used two local positional encodings for them.

In addition to 1D sequential positions, tables have structured rows and columns that contain two-dimensional and hierarchical information. 
Works such as TaPas, MATE, and TUTA, learned column/row embeddings based on column/row IDs and showed increased performance. However, considering hierarchical structures, as Figure~\ref{fig:realtable} (c) shows, column/row encoding results in limited representation capability. TUTA further devised explicit/implicit tree-based positional embeddings to jointly encode the spatial and hierarchical positions and showed significant effectiveness on generally structured tables. However, it has not proved helpful for downstream tasks that only involve flat and relational tables. 

\textbf{Numerical Encoding}
A large number of numerical values are distributed in tables and challenge BERT-based models to learn optimal representations since these methods simply tokenized and encoded numerical values in the same way as NL text. It corrupts the original recording structure of numbers into fragments and introduces difficulties in number representation~\cite{thawani2021representing}. However, understanding numerical values and their relationships is crucial for table content understanding, especially for tables in financial reports and scientific papers that involve a variety of numerical reasoning, such as aggregation, division, sorting, and change ratio. 

Explorations on learning better number representations have surged in the NLP field recently~\cite{thawani2021representing}, while few works attempted in tabular data. 
TaPas and MATE devised a unique rank embedding for column-wise number comparison, bringing improvements on answering comparatives or superlative questions. FLAP added extra feature encoding to indicate whether the text summary mentioned the value. TUTA distinguished numerical values from pure text via embedding over four discrete numerical features: magnitude, precision, the first digit, and the last digit. It is highly desirable to explore more numerical embedding methods in future works, e.g., considering arithmetical and statistical perspectives.

\textbf{Format Encoding} 
Formats contain valuable hints about table structures and data highlights, but only a few TaLMs considered them. E.g., TUTA learned format embeddings together with the transformer backbone to distinguish whether a cell has merge, border, formula, font bold, font color, and fill color.  

\subsubsection{Others} Apart from the typical encoding listed above, TaPas \cite{herzig2020tapas} designed "Previous Answer Embedding" for the conversational scenario, marking whether a tabular cell answered the last question. 

\begin{table}[t]
\begin{center}
\caption{ Encoder and decode architectures of TaLMs.}\label{tab:enc-dec}
\resizebox{.97\linewidth}{!}{%
\begin{tabular}{l  l }
\toprule
  \multirow{2}{*}{Encoder}  &  TaPas, TaBERT, TURL, TUTA, StruG, GraPPa, MATE, DoT, \\ 
   & ForTaP, TableFormer,  TABBIE, BRIDGE, etc.   \\ \midrule
  Encoder+Decoder  & KGPT, RPT, TaPEx, GAP, STTP, FLAP, UnifiedSKG, etc. \\ \midrule
  Decoder  & TableGPT, etc. \\
\bottomrule
\end{tabular}
}
\end{center}
\end{table}

\subsection{Encoder and Decoder Architecture}
Existing table pre-training models mainly inherit the architectures from language models such as BERT, GPT-2~\cite{radford2019gpt2}, and BART. Depending on the focused downstream tasks, these models adopted different components, i.e., encoders or decoders, as summarized in Table~\ref{tab:enc-dec}. Most of them adopted the \textbf{encoder} part of transformers similar to BERT, including TURL, StruG, TaPas, GraPPa, MATE, TUTA, ForTap, and TableFormer. Typically, a single encoder is applied on the sequential inputs constructed from tables and associated texts, if any, to learn the contextual representations of the inputs. Additional modules such as classification layers are applied to the encoder for downstream tasks. Some works even employed more than one encoder to capture the structural information of tables. TaBERT stacked column-wise self-attention layers on the row-wise encoder. TABBIE employed two encoders to encode the table row-wise and column-wise separately, then aggregated the representations obtained from both encoders. DoT~\cite{krichene2021dot} also used two encoders, one acting as a pruning model to select the most relevant tokens from the input and the other one used for performing specific tasks. Another 
branch of TaLMs used an \textbf{encoder-decoder} architecture to better enable sequence generation tasks such as table-to-text. KGPT used two alternative encoders, a GNN-based graph encoder and a transformer-based sequential encoder; each can be combined with a transformer decoder. RPT also adopted the encoder-decoder model, similar to a BERT model combined with a GPT model. TaPEx and UnifiedSKG implemented the encoder-decoder (text-to-text) model based on BART and T5, respectively, for downstream tasks such as text-to-SQL parsing, table-based QA, and table fact verification. STTP~\cite{xing2021sttp} focused on table-to-text generation by fine-tuning BART on table-structure-aware self-supervised tasks.
Instead of pre-training, TableGPT directly fine-tunes the GPT-2 \textbf{decoder} to take advantage of its contextual knowledge learned from linguistic corpora.  

\begin{table}[h]
\begin{center}
\caption{ Attention mechanisms of TaLMs.}\label{tab:attention}
\scalebox{0.78}{
\begin{tabular}{l l l l }
\toprule
\multirow{2}{*}{Dense}  & Self attention & TaPas, TaPEx, etc.  \\

  & Self attention with attention bias& TableFormer  \\
\hline
\multirow{5}{*}{Sparse} & Serial row/column attention layers & TaBERT  \\
& Parallel row/column attention layers  & TABBIE  \\
& Parallel row/column attention heads  & MATE \\
& Joint row/column attention  & TURL \\
& Joint bi-tree-based attention & TUTA \\
 & Joint layout-based graph attention & GTR \\
 & Joint knowledge-triple-based graph attention & KGPT \\
\bottomrule
\end{tabular}
}
\end{center}
\end{table}

\subsection{Structure-based Attention}

The attention mechanism is essential for TaLMs to compute contextual representations~\cite{vaswani2017attention}. Besides many of them that directly adopt self-attention, e.g., TaPas, StruG, GraPPa, and TaPEx, a series of structure-based attention mechanisms have been proposed to better leverage the tabular structure, as summarized in Table~\ref{tab:attention}. 

Self-attention may introduce a lot of irrelevant and even noisy contexts and have lots of unnecessary computations~\cite{wang2021tuta}, while table structures can be leveraged towards precise and efficient attention. In Table~\ref{tab:attention}, “dense” means all inputs in the table are visible in self-attention, while “sparse” means only parts of the table are visible.
TaBERT learned tabular representations serially by first producing row-wise encoding with a transformer and then column representation with vertical attention layers using row-wise encodings as inputs. 
TURL designed a restricted attention mechanism in which each token/entity attends to tokens in the same row/column.
TUTA proposed joint bi-tree-based attention, which took in both spatial and hierarchical information from tables. More specifically, for a structured table, TUTA defines cell coordinates and cell-to-cell distance from a bi-dimensional tree generated from the table. The bi-tree-based attention is restricted to attending tokens within the tree-based distance threshold. 
Cell embedding in TABBIE is an average of its row and column embeddings, where row/column embeddings are separately calculated by row/column transformers. 
MATE used different types of attention heads, i.e., row attention head and column attention head, which are restricted to attending tokens in the same row and column (query tokens can attend to all tokens).
GTR, which mainly focused on table retrieval, transformed a table into a tabular graph and used joint layout-based graph attention similar to the graph transformer to capture structural information. 
The graph transformer in KGPT enforced to use the structure (knowledge triple) as a hard constraint of attention, e.g.,  in the first encoder layer, each node was restricted to attend to tokens in its located knowledge triple. 
Tabnet~\cite{arik2019tabnet} applied a sequential attention mechanism to generate an interpretable feature selection mask during each decision step.
Rather than hard attention masking, TableFormer proposed to use soft attention biases when computing attention scores between two structural components. Structure-based attention not only improves model performance but potentially benefits model efficiency, and we will introduce it in Section~\ref{sec:efficiency}.

\subsection{Model Efficiency}
\label{sec:efficiency}
Most TaLMs are inefficient at dealing with long sequences due to the quadratic complexity of self-attention~\cite{vaswani2017attention,tay2020efficient}. Unfortunately, tables from webs and spreadsheets usually contain dozens of rows or columns, posing a significant challenge to the memory and computational efficiency~\cite{eisenschlos2021mate}. A naive way~\cite{herzig2020tapas,liu2021tapex} is truncating the input by a maximum sequence length, but it may lose critical information. So there emerge many other strategies:

\textbf{Input selection}~~One intuitive way is to filter the important part of the table before feeding them to the model. TaBERT extracted \textit{content snapshot}, the most relevant table rows to the NL sentence(s) regarding n-gram overlap ratio. Similarly, TaPas with intermediate pre-training~\cite{eisenschlos2020understanding} ranked columns by Jaccard coefficient between the NL and each column tokens. The model was twice as fast to train as TaPas while achieving similar performance. TUTA randomly sampled out $50\%$ text cells and $90\%$ numeric cells during pre-training since spreadsheets are usually large while limited semantics are introduced by similar data cells. Instead of using heuristics to prune inputs, DoT presented a model with two transformers: a pruning transformer selected top-$K$ tokens, and a task-specific transformer took them as inputs. 
The architecture was two times faster to train, while the memory bottleneck depended on the size of the pruning transformer. 
A small or medium-size pruning transformer was usually enough to achieve comparative performance with large-size TaPas, but falls behind on more challenging datasets like WTQ~\cite{pasupat2015compositional}.
While input selection is effective for tasks with table-text joint input like table QA and fact verification, it may fail for tasks that (1) all cells are required to be predicted, e.g., cell type classification; (2) table is the only input, e.g., table-to-text and formula prediction.

\textbf{Input splitting}~~An alternative way is to split a large table into multiple chunks and feed chunks separately into the model. TUTA split the table into chunks containing the same header row(s) and non-overlapped data rows in its downstream task, cell type classification~\cite{dong2019semantic}.  
For formula prediction, SpreadsheetCoder~\cite{chen2021spreadsheetcoder} (not a pre-training method but a typical case) split the target table by chunks with $N=3$ adjacent table rows/columns. The chunks were encoded by a BERT~\cite{devlin2018bert} encoder and then aggregated by convolutional layers. Splitting the input table allows encoding all table cells, while it costs more time due to multiple inferences of the encoder and is thus not used by most table pre-training methods. 

\textbf{Sparse attention}~~MATE leveraged the sparse attention of table encoding by an efficient implementation to reduce memory cost. In MATE, row and column attentions were designed for different attention heads, implemented following ETC~\cite{ainslie2020etc} by dividing the input into a global part $G$ attending to and from everything and a local part attending to $G$ and tokens within a radius $R$. The model scaled linearly concerning memory~($8,000$ tokens at most) and speed with this efficient attention implementation. Note that though sparse attention based on table structure is widely adopted by TaLMs, they mainly aimed to improve performance~(e.g., accuracy) instead of efficiency. Replacing these sparse attentions with efficient implementations can largely mitigate the memory issue: ForTaP used the TVM~\cite{chen2018tvm} framework to compile a CUDA kernel to implement sparse tree-attention in TUTA, and the maximum input sequence length was up to 8,000 tokens.

\section{Pre-training Objectives} \label{sec:pretraining objectives}

The pre-training objectives of TaLMs fall into two categories: Denoising autoencoder and task-specific objectives. Following the idea of Masked Language Modeling (MLM)~\cite{devlin2018bert}, many objectives adopt self-supervised labels for a TaLM to remove synthetic noises as autoencoder. Meanwhile, a variety of other pre-training objectives take inspirations from specific downstream tasks to design new supervisions.  The former apply self-supervised learning and denoise the table itself, and the latter build supervision according to external supervision signals or specific tasks.

\subsection{Denoising Autoencoder Objectives}
For denoising autoencoder objectives, TaLMs take partially corrupted inputs and recover the original ones. Most TaLMs applied token-level MLM on tabular sequences in the same way as NL sequences.
More advanced denoising objectives considered the table structure information such as cell and column.

\subsubsection{Token-level} Most pre-training models used token MLM ~\cite{devlin2018bert} by masking the input tokens at random and then predicting those masked tokens.  
TaPas, MATE, and TableFormer followed the standard MLM procedure to randomly mask 15\% of tokens. Larger ratio was taken by TURL (20\%) and TUTA (30\%) to make recovering more challenging.
Certain restrictions could also be applied on what tokens to mask. E.g., MLM used in TaBERT only masked tokens in external NL context, and MLM used in TURL and GraPPa masked both NL context and table headers.

\subsubsection{Cell-level} We list three kinds of cell-level objectives as follows.

\textbf{Masking and recovery}: 
A table cell could correspond to one or multiple token(s) in the tabular sequence. Slightly different masking strategies were designed.
    TUTA used a whole-cell masking strategy to capture relationships of neighboring cells.
    Cell Value Recovery (CVR) objective used in TaBERT applied the span-based prediction objective to deal with multiple value tokens.
    In TCN, each token represented a cell, and 10\% of table cells were masked for recovery from the set of cell values.
    
\textbf{Cell cloze}: 
TUTA sampled cell positions based on the bi-tree structure as candidate choices. At each sampled position, the model was encouraged to retrieve its corresponding cell string.

\textbf{Classifying cell corruption}:
TABBIE corrupted cells using two strategies, frequency-based and intra-table cell swapping.

\subsubsection{Column-level} Masked Column Prediction (MCP) was introduced by TaBERT to recover the names and data types of masked columns. GAP proposed to recover columns names using column values or the input utterance. Both of them assumed that tables were vertically-oriented and relational.

\begin{table*}[t]
\begin{center}
\caption{Downstream task evaluation for table pre-training. In this table, we try to cluster some similar tasks, e.g.,column type\&relation classification and a series of tasks of data preparation. We also merge sub-tasks like logic-to-text, into main tasks like table-to-text.}\label{tab:tasks}
\scalebox{0.79}{
\begin{tabular}{l l l}
    \toprule
    \textbf{Task} & \textbf{Description} & \textbf{Model} \\ 
    \midrule
    Table question answering & Given a table and a question in NL, output an answer. &  TaPas, TaBERT, StruG, GraPPa, TaPEx, GAP, MATE, \\
    ~ &  The question-answer pair should be supported by the table. & ForTaP, BRIDGE, TableFormer, UnifiedSKG \\
    \midrule
    Table fact verification & Verify whether a textual hypothesis holds based on the given
table. & MATE, TableFormer, UnifiedSKG\\
    \midrule
    Table-to-text & Generate textual description(s) from the given table. & KGPT, TableGPT, FLAP, STTP, UnifiedSKG \\
    \midrule
    Table type classification & Classify the table into different structural types.  & TUTA\\
    \midrule
    Cell type classification &  Identify
cell structural types in the table. & TUTA, ForTaP\\
    \midrule
    Column type &  Associate a column in a table with the KB type of entities it contains. & TURL, TABBIE, TCN\\
    \&relation classification & Associate a pair of columns in a table with the KB relation that & \\
    ~ & holds between each pair of entities in a given row of the columns. & ~ \\
    \midrule
    Table augmentation & Expand the table with additional data. & TURL, TABBIE\\
    \midrule
    Formula prediction &  Predict a spreadsheet formula for the target cell in the table. & ForTaP \\
    \midrule
    Entity linking &  Find phrases of text, called mentions, in cells and& TURL \\
    ~ & associate each with its referent entity. & ~ \\
    \midrule
    Table search &  Retrieve semantically relevant tables based on NL queries. & GTR \\
    \midrule
    Data preparation & Include six subtasks: data discovery, data validation, data filtering, & RPT \\
    ~ & data structuring, data enrichment, and data cleaning. & ~ \\
    \midrule
    Machine learning applications &  Various tasks using categorical or continuous features stored & Tabnet, VIME, TabTransformer \\
    ~ & in tabular format, e.g., the competitions held by Kaggle and KDD Cup. & ~ \\
    \bottomrule
    \end{tabular}
}
\end{center}
\end{table*}

\subsection{Task-specific Objectives}
\label{sec:downstreamobs}
To achieve SOTA performances on downstream task(s), denoising objectives might not be enough. 
Then task-specific objectives were proposed and proved to be highly effective.

\subsubsection{Objectives by Downstream Tasks}
In this section, representative objectives are grouped by tasks. Definitions of the downstream tasks can be found in Table~\ref{tab:tasks}.

\textbf{Table QA and semantic parsing:} 
There are a variety of works on table QA and semantic parsing, and we list some representative ones here.
A critical technical component of text-to-SQL is the alignment between text and tabular data. 
GraPPa designed an objective: given an NL sentence and table headers, predicting whether a column appears in the SQL query and what operation is triggered.
StruG used several grounding tasks of text-to-SQL as objectives, including selecting columns mentioned in sentences, finding cell values from sentences, and mapping column-value. 
TaPEx proposed to learn a novel neural SQL executor given a table and a synthesized executable SQL query. PReasM~\cite{yoran2021turning} synthesized at scale question-paragraph pairs that required different reasoning skills to enhance numerical reasoning abilities. GAP learned to compose complex SQL based on tabular data.

\textbf{Table fact verification:}
Entailment check is highly related with QA. \cite{eisenschlos2020understanding} used an intermediate pre-training objective of synthetic table fact checking targeting both real-world table fact verification and table QA. TaPEx, as described above, also showed benefits for table fact check.

\textbf{Entity linking:}
TURL proposed a Masked Entity Recovery (MER) objective by masking a certain percentage of input entity cells and then recovering the linked entity based on surrounding entity cells and table metadata. It helped the model capture the factual knowledge embedded in the table content and the associations between table metadata and table content.

\textbf{Table type classification and table search:}
TUTA provided each table with text segments and was pre-trained to retrieve the corresponding tables using text segments. 

\textbf{Numerical reasoning (formula-driven):}
ForTaP proposed to predict numerical reference and numerical calculation relationships, and aimed to benefit all related tasks involving table numerical reasoning, e.g., table QA and formula prediction.

\subsubsection{Objectives by Data Sources}
The above objectives are possible with human-created and machine-synthesized data on different sources of tables.

\textbf{Human-created:}
Human-created data usually show higher quality than web-crawled ones which might require careful prepossessing for their size, diversity and noises.
It is common to manually add extra labels for existing dataset. E.g., ToTTo, a well-labeled dataset for table-to-text with NL descriptions and corresponding web tables, was used by StruG for pre-training.
Also, human-created labels can be collected in smart ways. E.g., ForTaP extracted existing formulas from a large web-crawled spreadsheet corpus and extracted numerical reference and calculation relationships from them for pre-training. 
And large fine-grained labeled datasets were also used for pre-training, e.g., ToTTo, a well-labeled dataset for table-to-text with NL descriptions and corresponding web tables, was used by StruG for pre-training.

\textbf{Machine-synthesized:}
Synthesized data are more targeted and controllable, but require careful designs to ensure meaningfulness and diversity.
GraPPa proposed an SCFG (synchronous context-free grammar) and applied it to synthesize sentence-SQL pairs over tables.
\cite{eisenschlos2020understanding} created  counterfactual and synthetic statements for existing Wikipedia tables: For the counterfactual ones, it got tables and sentences from Wikipedia as positive examples and created minimally differing refuted examples; For the synthetic ones, it built table-dependent statements by synthesizing them from the pre-defined probabilistic CFG (context-free grammar).
TaPEx randomly selected tables from the training set of WIKITQ~\cite{pasupat2015compositional} and instantiated SQL templates to synthesize table-SQL pairs.

\section{Downstream Tasks}
\label{sec:task}
\begin{figure}[t]
    \begin{center}
    \includegraphics[width=3.4in]{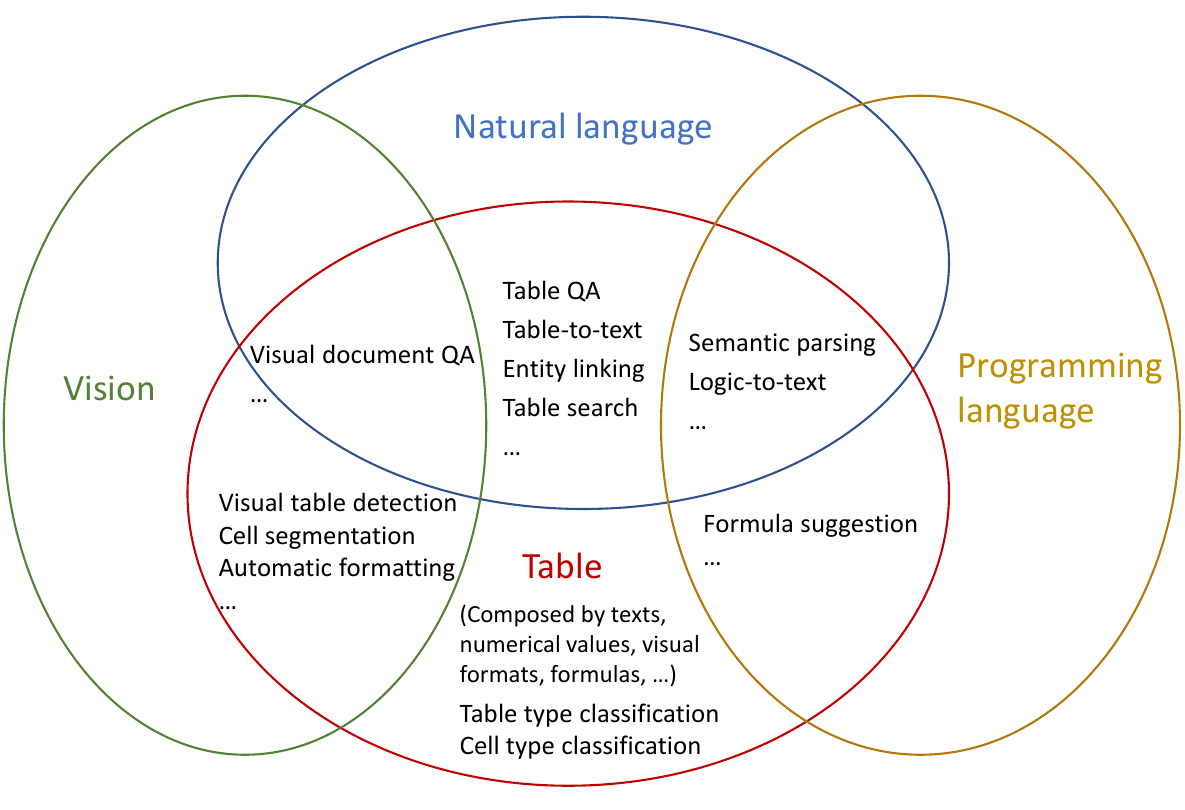}
    \end{center}
\caption{Downstream task categorization by domains. 
}
\label{fig:task}
\end{figure}

As shown in Figure~\ref{fig:task}, tasks of table understanding often have intersections with domains like NL, programming language, and computer vision, and thus prefer different capabilities of table modeling.
For example, table question answering is a prevalent cross-domain task that requires models to understand tables and NL questions jointly, and to enable robust reasoning over tables, semantic parsing becomes a widely-studied task of parsing NL questions to programming languages such as SQLs, logical forms, or python code. Question answering over visual documents is a cross-domain task of computer vision, tabular data, and NL. 
We think that classifying tasks by domains presents a fresh perspective to future works on multi-modal modeling, but it is not an absolutely strict or static categorization, e.g., table QA can involve programming languages via semantic parsing~\cite{yin2020tabert}, while it can also use end-to-end prediction~\cite{herzig2020tapas} without explicitly using a programming language. In addition to categorizing tasks by domains, tasks can also be categorized by their scenarios, e.g., data preparation represents a range of tasks for data preparation. The machine learning application covers various machine learning benchmarks or competitions where features and labels are stored in a tabular form. 

In Table~\ref{tab:tasks}, due to space limitation, we only list downstream tasks that have been evaluated by existing TaLMs. Nonetheless, it is highly desirable for future work to demonstrate the effectiveness of table pre-training on more tasks such as table format generation~\cite{dong2020neural}, data analysis~\cite{zhou2020table2analysis}, and table error detection~\cite{huang2018auto,zhang2021semantic}.

\section{Conclusion and Discussion}
\label{sec:conclusion}
This paper presents a comprehensive review of model architectures, objectives, and downstream tasks of table pre-training. 
Although the ``pre-training and fine-tuning'' paradigm has demonstrated its success on many table tasks, there still exist key challenges (opportunities) for future research:

\textbf{Combining Diverse Bi-dimensional Cell Features Effectively}
 Tables are arranged in two-dimension, including not only text but also quantities, visual formats, hyperlinks, and even spreadsheet formulas, it is non-trivial to learn high-level representations from such diverse and raw information. It particularly challenges existing language models that directly consume a flat sequence. Should tables be linearized like NL text? How to maintain and combine the structural, textual, formatting, and numerical information in a most effective way is still an open challenge.
 
\textbf{Universal Framework for Downstream Tasks}
Almost all TaLMs only focus on one or two downstream tasks (as shown in Table~\ref{tab:tasks}) so that they have sufficient flexibility on model designs (e.g., Section~\ref{sec:downstreamobs}) to achieve the best performance. But it is desirable to unify the advantages of existing methods and support various downstream tasks simultaneously like what BERT and GPT did in the NL domain. However, the diversity of table downstream tasks presents a significant challenge, e.g., entity linking and formula prediction and entity linking may need far different feature sets, sampling mechanisms, and model capabilities. It's a highly demanding direction to explore a universal framework by integrating the advantages of different TaLMs.

\textbf{Visualizing, Probing, and Comparing What Aspects TaLMs Learned from Table Pre-training}
To achieve a universal framework, we need to dig more deeper into what aspects of existing TaLMs learned from table pre-training.
Attention layers in transformers are often challenged for being opaque. In the NL domain, to uncover linguistic and world knowledge learned by pre-trained LMs, there exist attentive works on studying the outputs of pre-trained LMs on carefully designed input sentences~\cite{linzen2016assessing}, examining the internal vector representations of pre-trained LMs through methods such as probing tasks~\cite{adi2016fine,belinkov2017neural}, and 
visualizing attention maps of a pre-trained LMs~\cite{bahdanau2014neural,vig2019multiscale,rogers2020primer}. Later works even studied and probed attention layers head-by-head and found substantial syntactic information captured in BERT~\cite{clark2019does}. Some TaLMs claimed their abilities on table-text joint reasoning (TaPas, TaBERT, TaPEx, ...), table structure understanding (TUTA, TableFormer, ...), and numerical reasoning (ForTaP, ...), but there still lack attentive works on visualizing, probing, and comparing these intangible pre-trained TaLM models, leaving a large space to be exploited. 

\textbf{Consistency and Discrepancy Between LMs and TaLMs}
Recently, UnifiedSKG explored to directly fine-tune T5 on 21 datasets across 6 tasks. On the one hand, the frontier LM (T5) could easily achieve promising or even SOTA results on table tasks, demonstrating a strong relationship between text and tables, e.g., in linguistic and world knowledge aspects. On the other hand, without table-specific model design and pre-training, it still fell far behind TaLMs with table pre-training on WikiTQ (-8.2\%) and SQA (-12.1\%); even larger margins may exist on untested tasks, such as formula prediction. It shows the necessity of table-specific pre-training, e.g., in structural and reasoning aspects. 
(1) Is the initialization from advanced LMs necessary for table pre-training? Will TaLMs gradually ``forget'' the inherited knowledge from LMs during large-scale table pre-training?
(2) When zero-shot LMs are large enough (considering that GPT-3 has not been fully exploited on table tasks yet), do existing table pre-training strategies still have performance gains?
(3) During table pre-training, how to best inherit the knowledge from LMs while exploit table-specific capabilities? 

\textbf{Efficient Transformers for Big Tables}
Due to the limitation of the input length of existing large-scale pre-trained language models, they usually need to split tables into several parts~\cite{wang2020tuta} to be input to the model or only feed a specific part of a table (sometimes called "content snapshot") to the model according to the requirement of a specific task ~\cite{yin2020tabert,herzig2020tapas}. ForTaP and MATE are great trials but still have a limited maximum length (around 8,000 tokens) constraint that is not applicable to larger tables, such as a large proportion of spreadsheet tables. We wonder if there exists a general sampling method that is effective and general for various downstream tasks, and if there does not exist one, what we can do better in the pre-training stage.

\small
\bibliographystyle{named}
\bibliography{sample-sigconf}
\clearpage

\appendix
\clearpage

\begin{table*}[h]
\begin{center}
\caption{ Downstream task evaluation for table pre-training, a 2D view of Table~\ref{tab:tasks}.}\label{tab:tasks2}
\scalebox{0.69}{
\begin{tabular}{l  l l l l l l l l l l l l}
\hline

& Table & Table & Table & Table & Cell  & Column & Table & Formula & Entity & Table & Data & Machine\\
 & question & fact & -to-text & type & type & type \& relation & augmentation & prediction & linking & search & preparation & learning\\
& answering & verification & & classification & classification & classification & & & & & & applications \\
& Sec~\ref{sec:QA}  & Sec~\ref{sec:TFV} & Sec~\ref{sec:TTT} & Sec~\ref{sec:TTC} & Sec~\ref{sec:CTC} & Sec~\ref{sec:TI} & Sec~\ref{sec:TA} & Sec~\ref{sec:FP} & Sec~\ref{sec:TI} & Sec~\ref{sec:TS} &  Sec~\ref{sec:TDP} & Sec~\ref{sec:MLA} \\
\hline
TAPAS & yes &   &   &   &   &   &   &   &  \\
\hline
TaBERT & yes &   &  &    &   &   &   &   &   & \\
\hline
TURL &   &   &   & &   & yes & yes &   & yes & \\
\hline
TUTA &   &   &  & yes  &  yes &   &   &   &  \\
\hline
Tabnet &  &   &   &   &   &   &   &   &   &  & & Yes\\
\hline
VIME &  &   &   &   &   &   &   &   &   &  & & Yes\\
\hline
KGPT &  &  & yes &   &   &   &   &   &   & \\
\hline
RPT &  &   &  &   &   &   &   &   &  & & yes\\
\hline
StruG & yes &  &  &   &   &   &   &   &   &  \\
\hline
TabTransformer &  &   &   &   &   &   &   &   &   &  & & Yes\\
\hline
GraPPa & yes &  & &  &   &   &   &   &   &  \\
\hline
TABBIE & & &  &   &   &  yes & yes & &  \\
\hline
TaPEx & yes & &  &   &   &   &   &   &   &  \\
\hline
FORTAP & yes &  &  &  & yes   &   & & yes &  \\
\hline
MATE & yes & yes  &  &   &   &   &   &   &   &  \\
\hline
TableGPT &  &  & yes &   &   &   &   &   &   & \\
\hline
GTR &  &  &   &   &   &   &   &   &  & yes\\
\hline
TableFormer & yes & yes  &   &   &   &   &   &   &   & \\
\hline
FLAP & & & yes &  &   &   &   &   &   & \\
\hline
UnifiedSKG & yes & yes & yes &  &   &   &   &   &   & \\
\hline
\end{tabular}
}
\end{center}
\end{table*}

\section{Downstream Tasks and Datasets}

\subsection{Table Structure Understanding} 
Understanding the structure of tables is the premise of many other downstream tasks, e.g., question answering, table-to-text, and formula suggestion. Even the pre-training stage relies on the table type and header regions, e.g., in the data pre-processing stage of pre-training, TaBERT filters relational tables and "flattens" top headers if they cross multiple rows, and TUTA extracts tree-based positions 
based on header hierarchies. This section introduces two representative tasks of table structure understanding, table type classification and cell type classification. 

\subsubsection{Cell Type Classification} \label{sec:CTC}
Cell type classification is to identify
cell types in the table~\cite{dong2019semantic,koci2019deco,gol2019tabular}. Existing public datasets include deexcelerator (DeEx)~\cite{koci2019deco}, SAUS~\cite{gol2019tabular}, and CIUS~\cite{gol2019tabular}, where cells are categorize into metadata, notes, data, top attribute, left attribute, and derived. WebSheet~\cite{dong2019semantic} provides a fine-grained taxonomy of headers, including index, index name, and value name, but is not publicly available now. To better leverage cell-level information, ~\cite{LSTMbasedformat} encodes contextual (cell text) and stylistic (cell formats) features to embeddings via unsupervised training, and then trains an LSTM-based model for supervised cell classification. TUTA ~\cite{wang2020tuta} optimizes cell embedding layers, attention layers, and the cell classification head jointly via a transformer-based framework.

\subsubsection{Table Type Classification} \label{sec:TTC}
This task is to classify tables into different structural types~\cite{nishida2017understanding,ghasemi2018tabvec,eberius2015building}. However, most datasets are not publicly available, except \textsl{the July 2015 Common Crawl} (WCC) in \cite{ghasemi2018tabvec} where tables are categorized into five types: relational, entity, matrix, list, and non-data. To enable table-level classification, TUTA extends its pre-training backbone with a simple multi-layer classification head upon the special global [CLS] token of the table~\cite{wang2020tuta}.

\subsection{Table and Natural Language}

\subsubsection{Question Answering over Tables}  \label{sec:QA}
A general problem formulation of Table QA is defined as follows: given a table and a question in NL, output an answer, and the question-answer pair should be fully supported by the table.
Table QA has many benchmark datasets, mainly focusing on DB tables~\cite{wang2015building,yu2018spider,zhong2017seq2sql} and relational web tables~\cite{pasupat2015compositional,sun2016table}. Recently, it has been extended to QA~\cite{cheng2021hitab} over complex statistical report tables (questions are revised from real sentences that are meaningful, diverse, and natural rather than written from scratch),  QA~\cite{chen2021finqa,zhu2021tat} over financial tables, and jointly QA over tables and text~\cite{chen2020hybridqa,chen2021finqa,zhu2021tat}. In general, applying pre-training models to this task has three kinds of approaches: (1) TaBERT~\cite{yin2020tabert} feeds the embeddings outputted by the encoder to a logical forms generator and uses the generated logical forms to return answers by executing on tables; (2) TAPAS~\cite{herzig2020tapas} does not generate logical forms, and simply extends the encoder with classification heads for cell selection and operator selection; (3) TaPEx~\cite{liu2021tapex} uses an encoder-decoder architecture that decodes answers in a way like a sequence generation.

\subsubsection{Table-to-Text} \label{sec:TTT}
Table-to-text, i.e., generating textual descriptions from (semi-) structured tables, is another prevalent task that can benefit from table pre-training. 
There are many existing datasets for table-to-text ~\cite{liang2009learning,chen2008learning,wiseman2017challenges,novikova2016crowd,banik2013kbgen,lebret2016neural}. Recently, a large-scale controlled table-to-text dataset (ToTTo) is proposed for text generation over Wikipedia tables with annotated highlighted cells as conditions~\cite{parikh2020totto}. NumericNLG~\cite{suadaa2021towards} and SciGen~\cite{moosavi2021learning} are proposed to generate reasoning-aware paragraph-level descriptions for tables in scientific papers. LogicNLG~\cite{chen2020logical} is built for generating logical textual descriptions from Wikipedia tables. Logic2Text~\cite{chen2020logic2text} shares the same idea with LogicNLG to incorporate logical inference in table-to-text but contains annotated logical forms to enhance the generation fidelity. HiTab~\cite{cheng2021hitab} is the first to introduce hierarchical tables as the generation context, posing new challenges compared with previous datasets. Many prior approaches~\cite{gong2020tablegpt,kale2020text,chen2020logical,cheng2021hitab} on these datasets directly utilize pre-trained language models by first transforming the table into a textual sequence. A recent branch of works~\cite{xing2021sttp,flap2021} has focused on better table understanding by pre-training the transformer model on structure-aware self-supervised tasks.

\subsubsection{Table Fact Verification} \label{sec:TFV}
This is a binary entailment task to verify whether a textual hypothesis holds based on the given
table. TabFact~\cite{chen2019tabfact} is a big human-labeled dataset for table fact verification to evaluate the effectiveness of table-text joint modeling. Methods like TAPAS~\cite{eisenschlos2020understanding} and TaPEx~\cite{liu2021tapex} encode table and text together and outputs the truth value of entailment via a simple MLP-based binary classifier over the [CLS] token embeddings and a transformer-based decoder, respectively.

\subsubsection{Table Search by Natural Language} \label{sec:TS}

Table search by NL seeks to retrieve semantically relevant tables based on NL queries. WikiTables~\cite{zhang2018ad} is a benchmark dataset containing 60 queries and 3,120 candidate tables. WebQueryTable~\cite{sun2019content} contains 21,113
queries collected from search logs of a commercial search engine and 273,816 candidate tables. GTR~\cite{wang2021retrieving} encodes tables with a tabular graph transformer, contexts with a BERT-based model~\cite{devlin2018bert}, and queries with a FastText method~\cite{joulin2016bag}, and calculates the relevance score of the query-table and query-context matching by multi-layer perceptron (MLP).

\subsection{Table and Programming Language}

\subsubsection{Semantic Parsing Towards Table Question Answering}  \label{sec:SP}
Semantic parsing, usually seen as an important intermediate task of table question answering, is to parse natural language queries to SQLs (text-to-SQL), logical forms, or other kinds of programming languages that can return answers through executing on one or more target tables. Public datasets includes Spider~\cite{yu2018spider}, WikiSQL~\cite{zhong2017seq2sql}, etc. For example, StruG~\cite{deng2020structure}, GraPPa~\cite{yu2020grappa}, GAP ~\cite{shi2020learning} promote semantic parsing through pre-training on synthetic or human-labeled table-text data.

\subsubsection{Formula Prediction} \label{sec:FP}
Formula prediction is to predict a formula for the target cell in a table~\cite{chen2021spreadsheetcoder}. Since writing formulas from scratch could be time-consuming
and error-prone, it facilitates spreadsheet end-users by recommending formulas based on corresponding headers and other useful contexts. Enron~\cite{hermans2015enron} is a large and public dataset of Excel Spreadsheets containing over 17K spreadsheets with rich table structures and formula types. In FORTAP~\cite{cheng2021fortap}, based on the contextual embeddings of the target cell outputted by the pre-trained encoder, an LSTM-based decoder is adopted for formula generation.

\subsubsection{Formula Error Detection} \label{sec:FEP}
Formula error detection refers to the task of detecting spreadsheet formula errors. This task has real critical usage in the financial domain since formula errors can greatly degrade the quality of
spreadsheets and cause financial losses~\cite{powell2008critical}. Public datasets include CACheck~\cite{dou2016cacheck} and CUSTODES~\cite{cheung2016custodes}.

\subsection{Table and Vision}

\subsubsection{Image-based Table Detection and Cell Recognition}

Since image tables are not the main focus of this paper as claimed in Section~\ref{sec:pre-type}, we only list two most representative tasks for image tables here, image-based table detection and cell recognition. 
Image-based table detection is a task of detecting table regions in visual documents, and cell recognition further decomposes a whole table region into individual cell regions. Recent public datasets include ICDAR 2019~\cite{gao2019icdar}, TableBank~\cite{li2020tablebank},
PubTabNet~\cite{zhong2019publaynet}, SciTSR~\cite{chi2019complicated}, and FinTabNet~\cite{zheng2021global}. pre-training works on visual documents like LayoutLM~\cite{xu2020layoutlm}, DocFormer~\cite{appalaraju2021docformer}, and LAMBERT~\cite{garncarek2020lambert} can be potentially applied to these datasets. We will not expand this part due to the focus of this paper as stated in \ref{sec:pre-type}. 

\subsubsection{Table Format Generation}
Good-looking formatting is desirable for modern tables. Format generation refers to the task of generating formatting to better exhibit table structures and data relationships in a visual perspective. ~\cite{dong2020neural} provides three web-crawled datasets, SAUS, NCSE, and NSF, containing a total of 7,541 tables with high-quality table formatting.

\subsection{Table and Knowledge Graph}
\subsubsection{Table Interpretation}
\label{sec:TI}
Table interpretation is the task of converting Web tables into machine-understandable knowledge, including entity linking, column type identification, and relation extraction~\cite{bhagavatula2015tabel}. \textit{Entity linking} is the task of finding phrases of text, called mentions, in cells and associating each with its referent entity. \textit{Column type classification} is the task of associating a column in a table with the KB type of entities it contains. \textit{Relation extraction} is the task of associating a pair of columns in a table with the KB relation that holds between each pair of entities in a given row of the columns. 

T2Dv2 Gold Standard~\footnote{http://webdatacommons.org/webtables/goldstandardV2.html} contains 779 tables from various websites with manual annotations. 
SemTab challenge~\footnote{http://www.cs.ox.ac.uk/isg/challenges/sem-tab/} provides automatically generated tables with annotations from knowledge graph\cite{cutrona2020tough}. 
WikiTable corpus~\cite{bhagavatula2015tabel} contains entity linking annotation from hyperlinks, and TURL~\cite{deng2020turl} extends WikiTable corpus to the other tasks.

\subsubsection{Table Augmentation} \label{sec:TA}
Table augmentation is the task of expanding a table with additional data~\cite{zhang2017entitables}. Three sub-tasks are proposed for augmenting relational tables, including row population that suggests entity filings for table columns, cell filling that fills the cell values, and schema augmentation that recommends top headers to columns.
Since this task can be self-supervised, one can also adopt it as a pre-training objective using unlabeled data.

Knowledge graph augmentation is a task opposite to table augmentation. It enriches knowledge graphs with knowledge extracted from tabular data. Existing table pre-training works have not focused on this task.

\subsection{Table and Data Analysis}
\label{sec:TDA}
Tabular data is one of the important data formats in data analysis for business intelligence. There are two table analysis tasks -- table transformation and visualization generation. Table transformation aims at transforming tables with operators like group by, pivot, aggregation, and so on~\cite{yan2020auto}. Besides, some works ~\cite{zhou2020table2analysis} focus on creating analysis with transformation.
Visualization generation aims at recommendation visualization to make tabular data human-friendly and expressive. VizML~\cite{hu2019vizml} and Table2Chart~\cite{zhou2021table2charts} provide corresponding corpus.

\subsection{Table and Data Preparation} \label{sec:TDP}

Data preparation consists of a series of tasks. \cite{hameed2020data} classify them into six broader categories: data discovery, data validation, data structuring, data enrichment, data filtering, and data cleaning. 
RPT~\cite{tang2020rpt} improves a range of data preparation tasks through large pre-training with a transformer-based denoising autoencoder.
Here, we only introduce a representative data discovery task, spreadsheet table detection.  
\subsubsection{Spreadsheet Table Detection} \label{sec:STD}
Spreadsheets are flexible to use, but most tables on spreadsheets are not explicitly inserted by authors (by selecting the bounding box and clicking the "Insert Table" button in Excel), so spreadsheet table detection is essential to recognize table layouts and detect table boundaries. TableSense~\cite{dong2019tablesense} gives a public dataset containing 2,615 table regions from 1,645 spreadsheets ~\footnote{https://github.com/microsoft/TableSense}.

\subsection{Table and Machine Learning Application} \label{sec:MLA}
Since tabular data are often used to record categorical or continuous features and corresponding labels in machine learning applications (e.g., competitions held by Kaggle and KDD Cup such as product recommending and online advertising), this distinct scenario attracts research work such as VIME~\cite{yoon2020vime}, SAINT~\cite{somepalli2021saint}, TabTransformer~\cite{huang2020tabtransformer}, and Tabnet~\cite{arik2019tabnet}, to pre-train on unlabeled tabular data towards better featurization, feature selection, and contextual representation learning to achieve higher prediction accuracy in machine learning applications. Since there are pretty diverse machine learning applications, and most prediction tasks need complex reasoning over a specific set of domain-specific features, pre-training a general model has distinct challenges to handle those task diversity and reasoning complexity.
\subsection{Other Tasks}

There are still many related tasks and datasets that have not been evaluated by table existing pre-training methods, but they are also helpful to evaluate the effectiveness of pre-training models, such as table title generation~\cite{hancock2019generating}, pairwise table relation classification~\cite{fetahu2019tablenet},  tabular data synthesis~\cite{park2018data}, table error detection~\cite{huang2018auto}, etc.

\end{document}